%% file: main_WICCV.tex
\documentclass[10pt,twocolumn,letterpaper]{article}

\usepackage{iccv}              


\usepackage{titletoc}

\usepackage{color}
\usepackage{colortbl}
\usepackage{xcolor}
\usepackage{graphicx}
\usepackage{geometry}
\usepackage{float}
\usepackage{booktabs}
\usepackage{natbib} 
\usepackage{amsmath}
\usepackage{amssymb}
\usepackage{extsizes}
\usepackage{indentfirst}
\usepackage{longtable}
\usepackage{array} 
\usepackage{listings}
\usepackage{multirow}
\usepackage{comment}
\usepackage[utf8]{inputenc}
\usepackage[dvipsnames]{xcolor}

\usepackage{booktabs}
\usepackage{siunitx}
\usepackage{graphicx}

\usepackage{newunicodechar}
\newunicodechar{∙}{\textbullet}

\newcommand{\sofiia}[1]{\textcolor{blue}{[#1]}}

\input{math_commands.tex}

\definecolor{iccvblue}{rgb}{0.21,0.49,0.74}
\usepackage[pagebackref,breaklinks,colorlinks,allcolors=iccvblue]{hyperref}

\def\paperID{*****} 
\def\confName{ICCV}
\def\confYear{2025}

\title{Concept-Based Mechanistic Interpretability Using Structured Knowledge Graphs}

\author{
\textbf{
Sofiia Chorna\textsuperscript{\rm 1,2}, 
Kateryna Tarelkina\textsuperscript{\rm 1}, Eloïse Berthier \textsuperscript{\rm 1}, 
Gianni Franchi\textsuperscript{\rm 1}\footnotemark[1]} \\
U2IS, ENSTA, Institut Polytechnique de Paris\textsuperscript{\rm 1} \\
École Polytechnique Fédérale de Lausanne (EPFL)\textsuperscript{\rm 2}
}

\begin{document}
\maketitle

\input{sections/abstract}

\section{Introduction}
\input{sections/intro.tex}

\section{Related Work}
\input{sections/literature.tex}

\section{Global Concept-based Bias Analysis with \textsc{BAGEL}}
\label{sec:bagel_intro}
\input{sections/method.tex}

\section{Experimental Setup}
\input{sections/experiments.tex}

\section{Results}
\input{sections/results.tex}

\section{Conclusion}
\input{sections/conclusion.tex}
\clearpage

{
    \small
    \bibliographystyle{ieeenat_fullname}
    \bibliography{main}
}

\onecolumn
\appendix
\input{sections/appendix}

\end{document}

%% file: math_commands.tex
\usepackage{amsmath,amsfonts,bm}

\def\eqref#1{equation~\ref{#1}}

\def\1{\bm{1}}

\def\vomega{{\bm{\omega}}}

\def\vx{{\bm{x}}}

\DeclareMathAlphabet{\mathsfit}{\encodingdefault}{\sfdefault}{m}{sl}
\SetMathAlphabet{\mathsfit}{bold}{\encodingdefault}{\sfdefault}{bx}{n}



%% file: sections/abstract.tex
\begin{abstract}
While concept-based interpretability methods have traditionally focused on local explanations of neural network predictions, we propose a novel framework and interactive tool that extends these methods into the domain of mechanistic interpretability. Our approach enables a global dissection of model behavior by analyzing how high-level semantic attributes—referred to as \textit{concepts}—emerge, interact, and propagate through internal model components. Unlike prior work that isolates individual neurons or predictions, our framework systematically quantifies how semantic concepts are represented across layers, revealing latent circuits and information flow that underlie model decision-making.
A key innovation is our visualization platform that we named \textit{BAGEL} (for Bias Analysis with a Graph for global Explanation Layers), which presents these insights in a structured knowledge graph, allowing users to explore concept-class relationships, identify spurious correlations, and enhance model trustworthiness. Our framework is model-agnostic, scalable, and contributes to a deeper understanding of how deep learning models generalize—or fail to—in the presence of dataset biases. 
The demonstration is available at \url{https://knowledge-graph-ui-4a7cb5.gitlab.io/}.
\end{abstract}

%% file: sections/intro.tex
\begin{figure}[h]
    \centering
   \includegraphics[width=1.1\linewidth]{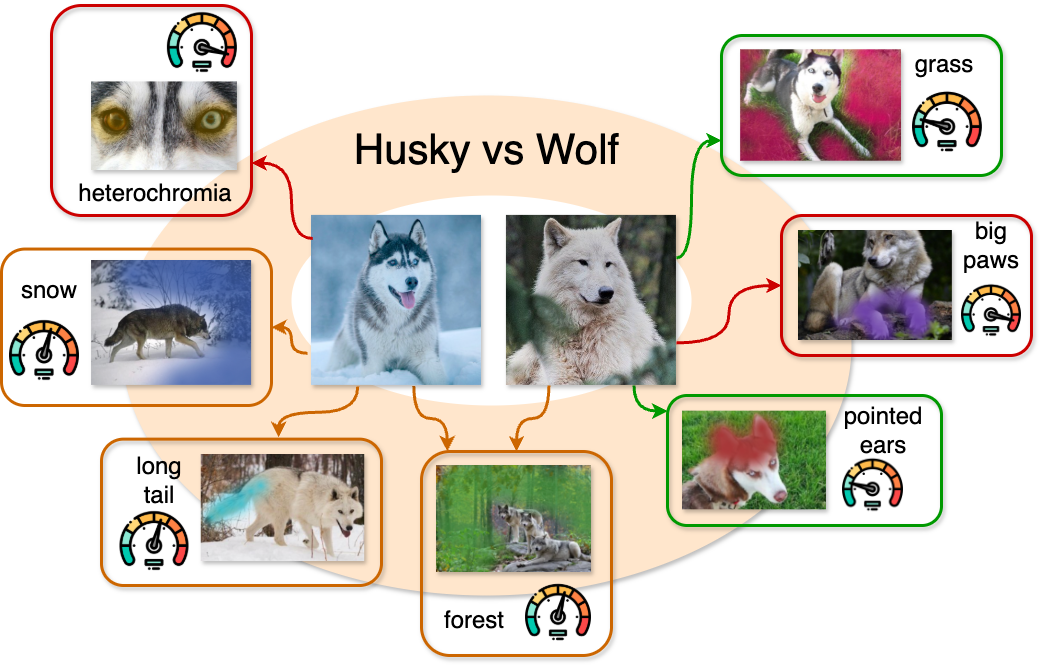}
    \caption{BAGEL. A conceptual network analysis approach supported by a knowledge graph. Central images show a husky (left) and a wolf (right); surrounding nodes illustrate corresponding key concepts such as eye color, fur, and natural habitats, with indicators showing bias significance.}
    \label{fig:bagel}
\end{figure}

Deep neural networks (DNNs) have demonstrated remarkable performance across a wide range of vision~\cite{krizhevsky2012imagenet, dosovitskiy2021imageworth16x16words}, language \cite{devlin2019bert}, and multimodal tasks ~\cite{radford2021learning}. Despite their success, these models remain notoriously difficult to interpret~\cite{arrieta2020explainable, kazmierczak2025explainability}. In safety-critical domains such as healthcare and autonomous driving, this lack of transparency can severely undermine trust and usability, particularly when users cannot understand the rationale behind model predictions.

Explainable Artificial Intelligence (XAI)~\cite{dwivedi2023explainable, saeed2023explainable} seeks to bridge this gap by developing tools and methods that render models more interpretable. Traditional white-box models—such as decision trees~\cite{decision_tree}, support vector machines~\cite{cortes_support-vector_1995}, and linear regression~\cite{linear_regression}—are inherently transparent but often sacrifice accuracy on complex tasks. In contrast, DNNs achieve superior performance but are widely regarded as black boxes.

Most XAI methods to date have concentrated on local explanations, focusing on individual predictions. However, such localized interpretations fall short in identifying broader patterns of bias and concept entanglement that arise from the training data and model architecture. This paper proposes a global, concept-based interpretability framework that addresses this limitation.

Mechanistic interpretability is a growing subfield of XAI that aims to reverse-engineer neural networks by uncovering how internal components—such as neurons, circuits, or attention heads—implement specific computations~\cite{olah2020zoom, nandaprogress}. Instead of merely observing model outputs, mechanistic interpretability seeks to identify the internal structures and pathways responsible for decision-making. This approach is particularly important for understanding generalization, detecting spurious correlations, and ensuring safety in high-stakes applications. Our proposed technique, \textbf{BAGEL}, contributes to mechanistic interpretability by modeling how semantic concepts are represented and interact within the network. By linking concept-level semantics to internal representations through a structured graph, our method sheds light on the mechanisms underlying the model's reasoning processes, thus bridging the gap between functional explanations and structural understanding.

We introduce a model-agnostic methodology to analyze how DNN classifiers leverage high-level, human-interpretable concepts across entire datasets. This methodology is based on a toolbox that aims to provide a global explanation of biases in DNNs, we named the toolbox \textit{BAGEL} for Bias Analysis with a Graph for global Explanation Layers.
In our framework, \textit{concepts} are defined as semantic attributes such as color, texture, object part, or scene context. These attributes serve as a bridge between human understanding and the opaque latent space learned by the model. For example, the classification of a ``zebra'' image might involve concepts like ``striped texture,'' ``savannah background,'' or ``quadruped shape.''

Our approach \textit{BAGEL} goes beyond traditional concept attribution by examining how concepts are distributed both in the dataset and the model’s internal representations. Specifically, we (1) analyze the presence of concept-based biases within the dataset, (2) model the alignment between these biases and those learned by the network, and (3) measure the similarity between dataset-driven and model-driven concept associations. This comparative analysis allows us to investigate whether the model relies on genuine features or spurious correlations, and (4) provide an interactive visualization tool that organizes concept-class relationships into a structured knowledge graph, enabling users to explore and compare dataset and model biases at scale.

%% file: sections/literature.tex
To contextualize our contributions, we organize related work into four categories: (1) concept-based ante-hoc methods, (2) concept-based post-hoc methods, (3) bias analysis approaches, and (4) neuron-level interpretation methods such as Network Dissection. This taxonomy helps delineate the evolution of interpretability tools and highlights the specific gap addressed by our approach.

\subsection{Concept-Based Ante-Hoc Methods}

Concept-based explanations aim to map the internal representations of deep learning models to human-understandable concepts. Ante-hoc approaches incorporate this mapping during model training. One prominent framework is the Concept Bottleneck Model (CBM)~\cite{pmlr-v119-koh20a}, introduced by Koh \etal, where a deep neural network is decomposed into two modules: a concept predictor and a label predictor. Both modules are trained using annotated concept labels in the standard supervised setting.

Several strategies exist for extracting concepts in CBMs. For instance, \cite{bennetot_greybox_2022} employs semantic segmentation, while \cite{diaz2022explainable} relies on object detection. Various supervised extensions have been proposed~\cite{mahinpei2021promises,havasi2022addressing}, yet they all depend on the availability of concept annotations, which can be costly or impractical in many settings.

Recent advances in Vision-Language Models (VLMs), such as CLIP~\cite{radford_learning_2021}, have enabled zero-shot concept understanding, sparking interest in unsupervised CBMs~\cite{oikarinen_label-free_2023, srivastava2024vlg, yang2023language, kazmierczak_clip-qda:_2023}. These methods leverage the joint text-image embedding space of CLIP to estimate concept relevance without requiring manual annotations. While these models present promising directions for built-in interpretability, they often rely on predefined or static concept sets. Our approach builds upon this line of work by analyzing the alignment between model-internal representations and dynamically derived, knowledge-driven concept structures.

\subsection{Concept-Based Post-Hoc Methods}

Post-hoc concept-based methods seek to interpret a trained model by identifying directions in its latent space corresponding to semantic concepts. A foundational method in this space is Concept Activation Vectors (CAVs)~\cite{kim_interpretability_2018}, which estimate concept sensitivity by training a linear classifier on internal activations of examples with and without the concept. 

Numerous extensions have improved the original formulation~\cite{ghorbani2019interpretation, ghorbani2019towards}. Bai \etal~\cite{bai_concept_2024} propose a nonlinear extension using kernel methods to overcome the limitations of linear separability. Tetkova \etal~\cite{tetkova_knowledge_2024} introduce knowledge graphs (e.g., WordNet, Wikidata) to automatically collect concept examples and guide the definition of semantically grounded CAVs. Their work also quantifies alignment between CAV directions and the structure of semantic graphs, showing that semantically similar concepts exhibit more aligned CAVs.

Other works integrate concept learning with dictionary methods. For instance, Fel \etal~\cite{fel2025archetypal, fel2023holistic} combine CAVs with sparse dictionary learning to interpret the latent space better, while CRAFT~\cite{fel2023craft} applies non-negative matrix factorization and recursive extraction to explain predictions via structured concepts.

Despite their effectiveness, CAV-based methods suffer from several limitations: they require curated concept examples, are sensitive to the choice of layer, and often assume linearity. Furthermore, CAVs typically focus on localized visualization rather than providing a unified comparison of concept representations across data and models. Our method addresses these gaps by offering a model-wide view of concept distributions and their alignment with external semantic priors.

\subsection{Bias Analysis Approaches}

Bias in machine learning systems can arise from dataset imbalances or model-specific inductive biases. Early works~\cite{torralba2011unbiased, khosla2012undoing} demonstrated how overrepresented visual contexts in datasets can lead to spurious correlations, compromising generalization.

On the model side, saliency-based~\cite{simonyan2013deep} and perturbation-based~\cite{fong2017interpretable} interpretability methods have been used to reveal biased behaviors. However, these approaches often struggle to connect observed biases with semantically meaningful or causal explanations. Moreover, their robustness has been questioned~\cite{adebayo2018sanity}, especially under randomized or adversarial settings.

Some recent works focus on bias mitigation~\cite{singh2020don}, but fewer offer tools for tracing how biases propagate through a model. Our approach contributes to this line of research by enabling the analysis of concept distributions across the network, allowing bias to be diagnosed not at the pixel level but through high-level semantic representations.

\subsection{Neuron-Level Interpretation and Network Dissection}

Understanding deep networks by inspecting the behavior of individual neurons has a long history~\cite{erhan2009visualizing, zeiler2014visualizing, olah2017feature}. This led to the development of Network Dissection~\cite{netdissect2017, bau2020understanding}, a technique that quantifies the alignment between neuron activations and human-interpretable concepts by computing the Intersection over Union (IoU) between activation maps and pixel-level segmentation annotations.

Although Network Dissection provides rich insights into how concepts are localized within individual neurons, it relies heavily on manually annotated datasets, which limits scalability and generalization. Extensions to LLMs~\cite{hernandez2021natural} show the broader applicability of the neuron-level paradigm but face similar limitations in requiring labeled data.

To address this, CLIP-DISSECT~\cite{oikarinenclip} leverages the zero-shot capabilities of VLMs to perform dissection without human annotations. By decoupling the concept set from the probing dataset, any network can be analyzed using arbitrary text corpora and image datasets, with CLIP-based similarity measures linking neurons to concepts. DISCOVER~\cite{panousis2023discover} further extends this idea to vision transformers by proposing methods tailored to their more complex architectures.

In our work, we adopt a similar philosophy but extend the analysis to all layers of the network. To facilitate comparison across layers and architectures, we apply global average pooling to convolutional units, effectively summarizing each unit's activation as a scalar. While this sacrifices spatial localization, it enables a unified treatment of both convolutional and fully connected layers, providing a broader view of concept alignment throughout the model layers.

%% file: sections/method.tex
In this section, we introduce \textsc{BAGEL}, a novel toolbox designed to provide a global understanding of the biases learned by deep neural networks (DNNs). Unlike traditional instance-level interpretability tools, \textsc{BAGEL} aims to capture how concept-level representations evolve across the network architecture and relate them to biases present in the training dataset. Our goal is to deliver a global explanation of a DNN’s internal representations, making biases both measurable and interpretable at the scale of entire datasets and layers.

The section is structured as follows: we begin by introducing the background in Section \ref{sec:Background}, followed by a discussion of dataset bias in Section \ref{sec:Dataset}. We then present our approach to modeling bias in Section \ref{sec:concept_prediction_using_embeddings}, and describe how to measure bias similarity in Section \ref{sec:dataset_model_bias_comparison}. This part is particularly important, as our objective is to assess the relationship between model and dataset biases. Finally, in Section \ref{sec:Graph}, we present the Knowledge Graph generated by BAGEL, which synthesizes all the gathered information.

\subsection{Background and Notations}
\label{sec:Background}
Let \( f \) be a DNN, viewed as a function of input data~\( \vx \) and trained weights \( \vomega \). The network’s output for an input \( \vx \) is denoted \( f_{\vomega}(\vx) \). We assume \( f \) is composed of \( L \) layers, such that:

\begin{equation}
f_{\vomega} = f_{\vomega^L} \circ \ldots \circ f_{\vomega^1},
\end{equation}
where \( \vomega^\ell \) are the weights of layer $\ell \in \{1,...,L \}$. We denote the output of layer \( \ell \) as \( f_{\ell}(\vx) \). Each layer may output a vector (for a fully connected layer) or a tensor of feature maps (for a convolutional layer). For simplicity, we restrict our analysis to these two layer types.

\paragraph{Unit-Level Representation.} Inspired by Network Dissection~\cite{netdissect2017}, we define a \emph{unit} as an individual neuron in the network, typically a single channel in a convolutional layer. The activation of unit \( u \) at spatial position \( p \) in layer \( \ell \) for input \( \vx \) is:
\begin{equation}\label{eq/unit1}
a_{\ell,u}(\vx, p) = \sigma\left( (W_{\ell,u} * f_{\ell-1}(\vx))(p) + b_{\ell,u} \right) ,
\end{equation}
where \( W_{\ell,u} \) is the convolutional kernel, \( b_{\ell,u} \) the bias, \( * \) denotes convolution, and \( \sigma \) is a non-linear activation (typically ReLU). This spatially varying activation allows concept localization.

\paragraph{Global Activation via GAP.} To analyze units across all layers uniformly (including fully connected ones), we apply Global Average Pooling (GAP) to the output of each convolutional layer:
\begin{equation}\label{eq/unit3}
\tilde{a}_{\ell,u}(\vx) = \frac{1}{|P|} \sum_{p \in P} a_{\ell,u}(\vx, p),
\end{equation}
where \( P \) is the set of spatial positions. This reduces each unit's response to a scalar, enabling a consistent representation:
\begin{equation}\label{eq/unit3}
\phi_{\ell}(\vx) = \begin{bmatrix}
\tilde{a}_{\ell,1}(\vx), \ldots, \tilde{a}_{\ell,U_{\ell}}(\vx)
\end{bmatrix} \in \mathbb{R}^{U_{\ell}} ,
\end{equation}
where \( U_{\ell} \) is the number of units in layer \( \ell \).

\subsection{Dataset Bias Quantification}\label{sec:Dataset}

Let \( \mathcal{D} \) be dataset composed of image $\vx$  and label \( y \in \{y_1, \ldots, y_I\} \) ($y_I$ is the number of classes and $y_i$ is one class). Inspired by CAV and CBM approaches, we characterize each image \( \vx \) by a set of visual concepts \( \{c_k\}_{k=1}^K \), such as colors, textures, objects, or contexts.

Dataset bias arises when certain concepts are disproportionately associated with particular classes. For instance, if the concept ``snow'' appears predominantly in the images of the class ``husky'', the network might spuriously associate snow with the husky class. We quantify such biases via the empirical conditional probability:
\begin{equation}
p_{\text{dataset}}(c_k \mid y_i) = \frac{N_{y_i,c_k}}{N_{y_i}} ,
\end{equation}
where \( N_{y_i,c_k} \) is the number of images in class \( y_i \) labeled with concept \( c_k \), and \( N_{y_i} \) is the total number of samples in class \( y_i \). To determine concept presence in an unsupervised manner, we use CLIP \cite{radford2021learning} in a zero-shot classification setting, where the target ``class'' is the concept name.

\subsection{Concept Prediction using Intermediate Layer Embeddings}
\label{sec:concept_prediction_using_embeddings}

We seek to understand whether concept-related information is captured by a given layer of the DNN. For each layer \( \ell \), we use the corresponding feature vector \( \phi_{\ell}(\vx) \) (define in eq. \ref{eq/unit3}) and train a concept classifier \( g_{\ell}^{(k)} \) to predict the presence of concept \( c_k \) from the layer's output.

Let \( y^k \in \{0,1\} \) denote the ground-truth presence of concept \( c_k \) in image \( \vx \). Then the classifier \( g_{\ell}^{(k)} \) estimates the probability:
\begin{equation}
p_k(\vx) = g_{\ell}^{(k)}(\phi_{\ell}(\vx)) .
\end{equation}
Each classifier \( g_{\ell}^{(k)} \) is a simple logistic regression trained to distinguish the presence or absence of the concept. Once trained, these classifiers let us estimate the prevalence of a concept within a class \( y_i \), at layer~$\ell$, by averaging the predicted probabilities across the images of the class:

\begin{equation}
p_{\text{model}}^{(\ell)}(c_k \mid y_i) = \frac{1}{N_{y_i}} \sum_{\vx \mbox{ of class } y_i} p_k(\vx) .
\end{equation}

This procedure allows \textsc{BAGEL} to produce a global concept-layer distribution, interpretable across both network and dataset scales.

\subsection{Dataset vs. Model Bias Comparison}
\label{sec:dataset_model_bias_comparison}

Dataset biases reflect the inherent skew in concept-class association within the training data, often arising from imbalanced sampling or human annotation tendencies. Model bias as a proxy for layer \(\ell\) captures how a DNN interprets these associations during training. Depending on its architecture, training, and regularization techniques, the model can reproduce dataset bias, reduce it by filtering spurious correlations, or strengthen it by overemphasizing certain patterns.

To evaluate whether a DNN replicates or mitigates dataset biases, we compare \( p_{\text{dataset}}(c_k \mid y_i) \) with \( p_{\text{model}}^{(\ell)}(c_k \mid y_i) \) for each layer \( \ell \), class \( y_i \), and concept~$c_k$. We employ two metrics to compare these probabilities.

\paragraph{Weighted F1-Score.} The score quantifies the alignment between the binarized concept distribution of the dataset and those predicted by the model. To enable this comparison, both the dataset and model probabilities are binarized using a threshold \(\tau\):
\begin{align*}
\hat{y}_k^{\text{dataset}} &= \mathbb{I}(p_{\text{dataset}}(c_k | y_i) \geq \tau) \\
\hat{y}_k^{\text{model}} &= \mathbb{I}(p_{\text{model}}^{(\ell)}(c_k | y_i) \geq \tau),
\end{align*}
where \( \mathbb{I} \) is the indicator function. The threshold~$\tau$ is set to 0.5 by default but can be adjusted in our toolbox \textsc{BAGEL}.

The weighted F1-score is computed as:
\begin{equation*}
    \text{F1}_{\text{weighted}} = \sum_{y_i} w_i \cdot \frac{2 \cdot \text{Precision}(y_i) \cdot \text{Recall}(y_i)}{\text{Precision}(y_i) + \text{Recall}(y_i)}, \label{eq:f1_weighted}
\end{equation*}
where \( w_i \) is the fraction of samples belonging to class \(i\) in the ground truth labels.

\paragraph{Jensen-Shannon (JS) Divergence.} JS divergence quantifies distributional differences between dataset and model probabilities, capturing subtle shifts in concept-class associations. It is defined as:
\begin{equation*}
    \text{JS} = \sqrt{\frac{1}{2} \text{KL}(p_{\text{dataset}} \parallel m) + \frac{1}{2} \text{KL}(p_{\text{model}}^{(\ell)} \parallel m)}, \label{eq:js_div}
\end{equation*}
where \( m = \frac{1}{2}(p_{\text{dataset}} + p_{\text{model}}^{(\ell)}) \), and KL is the Kullback-Leibler divergence. This metric is symmetric, bounded and measures how closely the model’s probabilities match the dataset’s, with lower values indicating better alignment.

\begin{figure*}[h]
    \centering
    \includegraphics[width=1.1\linewidth]{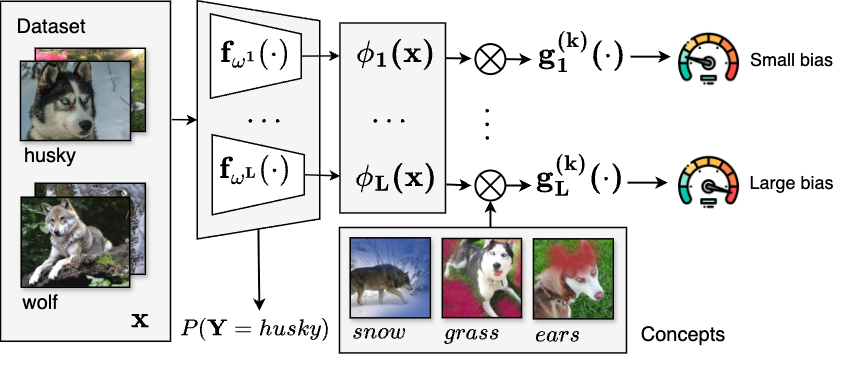}
    \caption{\textbf{Overview of the proposed \textsc{BAGEL} framework.} The distribution of the concepts in the embeddings across DNN layers is compared to the original concept distribution in the dataset. A knowledge graph visualizes the alignment between these distributions to enhance the exploration and understanding of biases.}
    \label{fig:method-overview}
\end{figure*}

\subsection{Knowledge Graph Representation}
\label{sec:Graph}

To visualize and analyze the interplay between dataset bias and model-internal concept activations across layers, we employ an interactive \textit{Knowledge Graph} (KG) representation \cite{Hogan_2021}. This KG captures probabilistic relationships between classes and semantic concepts extracted from both the dataset and the model's internal representations.

A Knowledge Graph is formally defined as a directed labeled graph:
\[
\mathcal{G} = (\mathcal{V}, \mathcal{E}, \mathcal{W}),
\]
where:
\begin{itemize}
    \item \(\mathcal{V} = \mathcal{Y} \cup \mathcal{C}\) is the set of nodes, consisting of class nodes \(\mathcal{Y} = \{y_1, \ldots, y_I\}\) \ and concept nodes \(\mathcal{C} = \{c_1, \dots, c_K\}\), 
    \item \(\mathcal{E} \subseteq \mathcal{Y} \times \mathcal{C}\) is the set of directed edges from classes to concepts,
    \item \(\mathcal{W} : \mathcal{E} \rightarrow [0,1]^L\) assigns a weight vector to each edge, where \(L\) is the number of layers in the model and each component \(w^{(\ell)}(e_{i,k})\) corresponds to the probability that the concept \(c_k\) is detected in the model’s layer \(\ell\) given the class \(y_i\).
\end{itemize}

We distinguish between two types of probabilities:
\begin{itemize}
    \item Dataset-based probability $p_{\text{dataset}}(c_k | y_i)$, 
    estimated empirically from the dataset,
    \item Model-based probability $p_{\text{model}}^{(\ell)}(c_k | y_i)$, 
    computed from the feature maps at layer \(\ell\).
\end{itemize}

An edge \(e_{i,k} = (y_i, c_k)\) is included in \(\mathcal{E}\) if at least one of these probabilities exceeds a predefined threshold \(\tau\), i.e., if:
\[
p_{\text{dataset}}(c_k | y_i) \geq \tau \quad \text{or} \quad \exists \ell \in [1, L], \;p_{\text{model}}^{(\ell)}(c_k | y_i) \geq \tau.
\]

\paragraph{Edge Semantics and Visualization.}
In our graphical interface, the KG is rendered with visual encodings to highlight bias overlap and divergence:
\begin{itemize}
    \item \colorbox{Green}{\textbf{Green edges:}}  \(p_{\text{dataset}}(c_k | y_i) \geq \tau\) and \(p_{\text{model}}^{(\ell)}(c_k | y_i) \geq \tau\) for some \(\ell\). Represents overlap between dataset and model biases.
    \item \colorbox{Blue}{\textbf{\textcolor{white}{Blue edges:}}} \(p_{\text{dataset}}(c_k | y_i) \geq \tau\) but \(p_{\text{model}}^{(\ell)}(c_k | y_i) < \tau\) for some \(\ell\). Indicates a dataset bias not learned by the model.
    \item \colorbox{Red}{\textbf{Red edges:}} \(p_{\text{dataset}}(c_k | y_i) < \tau\) but \(p_{\text{model}}^{(\ell)}(c_k | y_i) \geq \tau\) for some \(\ell\). Indicates model-specific bias not present in the dataset.
    \item \colorbox{lightgray}{\textbf{Light gray edges:}} Both probabilities are below the threshold; the concept is weakly associated or irrelevant.
\end{itemize}

The edge width is proportional to the highest model-based concept probability across all layers:
\[
\text{width}(e_{i,j}) \propto \max_{\ell \in [1,L]} p_\text{model}^{(\ell)}(n_j \mid c_i).
\]
Concept nodes are colored based on their semantic category (e.g., \textit{texture}, \textit{part}, \textit{material}) to facilitate visual grouping.

\paragraph{Objective.} 
This KG allows us to explore the relationship between dataset and model biases, particularly their overlap. By navigating through different layers and observing the dominant edge types, we assess whether the model's classification decisions rely on dataset-driven concepts or diverge toward independent, possibly spurious features.

\paragraph{Example.} 
Figure~\ref{fig:showcase} illustrates a KG for a single model layer. Orange nodes correspond to classes, and other nodes represent concepts. The color and thickness of the edges provide insight into the bias alignment and its strength.


\paragraph{Interactivity.}
The graph can be explored interactively via our GUI, where users can set the threshold \(\tau\), navigate across layers, and focus on specific classes or concept categories. This enables deeper insights into how the model learns and generalizes and whether it captures meaningful or misleading semantic structures from data.

In summary, \textsc{BAGEL} provides a novel, scalable framework for analyzing how DNNs encode, preserve, or distort dataset-level concept biases. The method supports a global explanation of model behavior and can help uncover spurious correlations at the network level. The BAGEL framework is depicted in Figure~\ref{fig:method-overview}.

%% file: sections/experiments.tex
In this section, we describe the experimental framework for assessing our proposed method, \textit{BAGEL}, on a diverse set of DNNs and datasets. Our experiments aim to assess its ability to detect and visualize  concept-based biases, quantify bias alignment between datasets and models, and provide quantitative and qualitative analyses. We explore a range of architectures to demonstrate method robustness.

\paragraph{Models.}
The following models were used in the comparison: AlexNet~\cite{krizhevsky2012imagenet}, ResNet18~\cite{he2016deep}, DenseNet121 and DenseNet169~\cite{huang2017densely}, ResNeXt101~\cite{xie2017aggregated}, VGG16~\cite{simonyan2014very}, EfficientNet-B0~\cite{tan2019efficientnet}, and GoogLeNet~\cite{szegedy2015going}. These models were chosen to span a range of architectural complexities, from shallow (e.g., AlexNet) to deep (e.g., ResNeXt101). They were fine-tuned on each dataset using hyperparameters detailed in Supp. ~\ref{appndx:training_details}.

\paragraph{Datasets.} We experimented with both concept-annotated and non-annotated datasets:
\begin{itemize}
    \item \textbf{Husky vs. Wolf} and \textbf{KitFox vs. RedFox}: binary classification subsets derived from ImageNet ~\cite{imagenet_cvpr09},
    \item \textbf{MonuMAI} \cite{monumai_2021}: architectural style classification (e.g., baroque, gothic) with annotated concepts (e.g., ``rounded arch'').
    \item \textbf{Cats/Dogs}: a dataset derived from \cite{kazmierczak_clip-qda:_2023} with an introduced color bias (white dogs, black cats).
    \item \textbf{Derm7pt} \cite{derm7pt_2019}: a medical dataset for skin lesion classification annotated with seven dermatological concepts (e.g., ``pigment network'').
\end{itemize}

For non-annotated datasets (Cats/Dogs, Husky vs. Wolf, KitFox vs. RedFox), we generated concepts using CLIP \cite{radford2021learning}. The list of concepts per dataset is provided in Supp. \ref{appndx:set_of_concepts}.

\paragraph{Metrics.} We measure concept detection and bias alignment using: (1) \textit{weighted F1-score}, which accounts for class imbalance in concept prevalence, and (2)\textit{ Jensen-Shannon (JS) divergence}, which quantifies the distributional differences between the dataset and model biases, as described in Section \ref{sec:dataset_model_bias_comparison}

\paragraph{Implementation.} Concept prediction was implemented using logistic regression classifiers trained on features extracted from DNNs layers via forward hooks (see Section \ref{sec:concept_prediction_using_embeddings}). Each logistic regression uses log loss (binary cross entropy) as a loss function and is optimized using the \textit{liblinear} solver with a maximum of 1000 iterations. To handle imbalanced concept prevalence, we apply balanced class weights. The regularization parameter \(C\) is tuned using grid search over ${0.01, 0.1, 1.0, 10.0}$ using 5-fold cross-validation, selecting the value that maximizes average precision (typically $C=0.1$ for most layers).

The choice of logistic regression is motivated by its interpretability: it provides a straightforward mapping from features to concept probabilities. However, we note that it assumes a linear relation between features and concepts, which may limit its ability to capture more complex patterns.

%% file: sections/results.tex
\subsection{Qualitative results}
To illustrate BAGEL's ability to detect and visualize biases, we present a qualitative analysis of concept-class relations using interactive KG, as detailed in Section \ref{sec:Graph}.

\begin{figure}[htpb]
\centering
\includegraphics[width=0.91\linewidth]{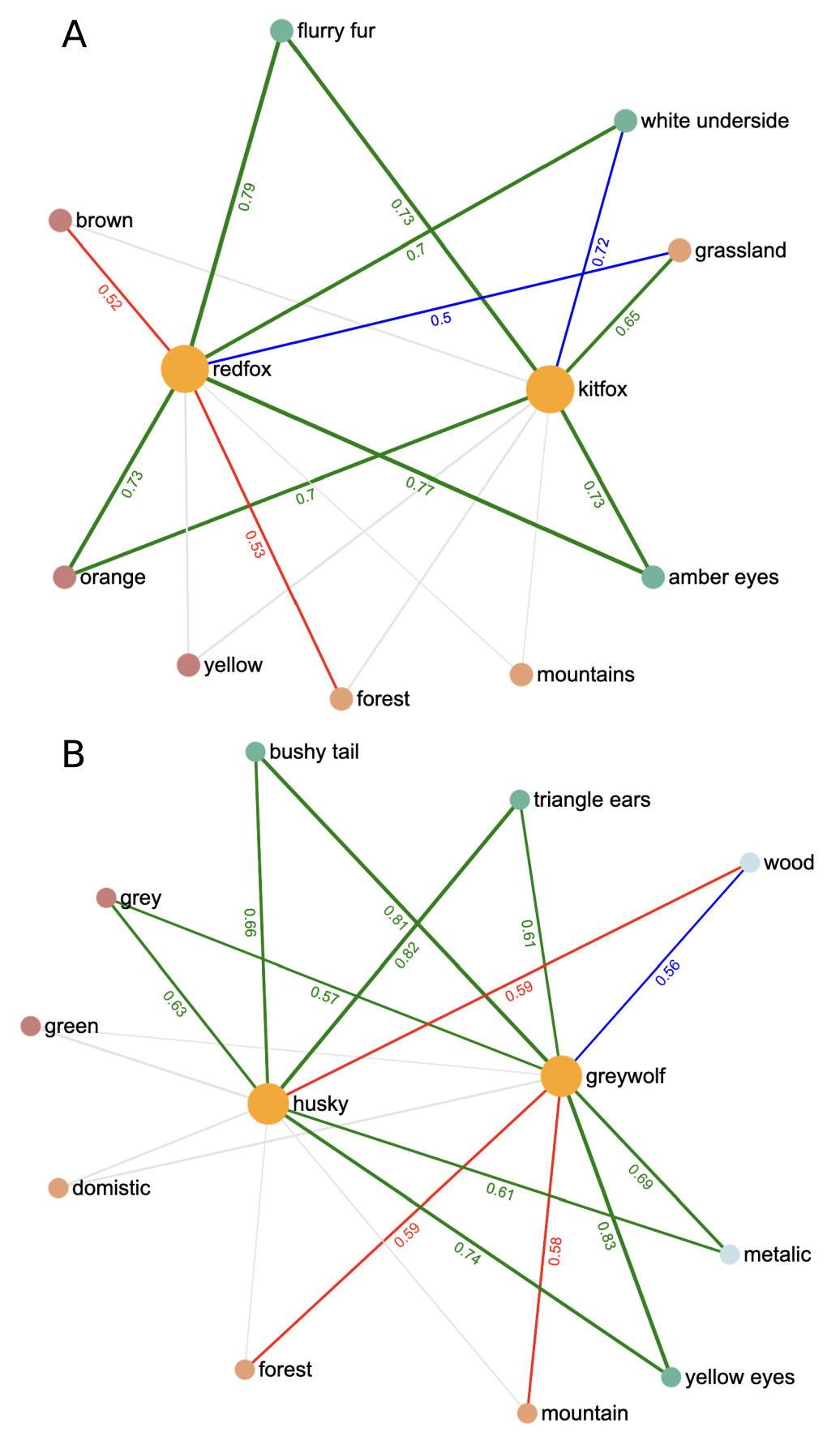}
\caption{\textbf{Examples of found biases using BAGEL}. \textcolor{orange}{ \Large\textbullet} Orange nodes are classes, colored nodes are concepts, edge colors show bias sources: \colorbox{Green}{green} (overlapping), \colorbox{Blue}{\textcolor{white}{blue}} (dataset-specific), \colorbox{Red}{red} (model-specific). \textbf{a)} ResNet50 (layer 3) on Redfox vs. Kitfox; \textbf{b)} DenseNet121 (denseblock4) on Husky vs. Wolf.}
\label{fig:showcase}
\end{figure}

Figure~\ref{fig:showcase}a) shows the KG generated for the third residual layer of ResNet50 on the Redfox vs.~Kitfox dataset. Predominantly green edges indicate aligned dataset-model biases for most concepts. However, the blue edges for the ``grassland'' and ``white underside'' concepts reflect data bias not fully learned by the model, and the red edges for the ``brown'' and ``forest'' concepts suggest model-specific biases absent from the dataset. It leads to the suggestion that the model may overemphasize color and environmental features irrelevant to the true class distinctions, possibly due to spurious correlations learned during the training. Figure ~\ref{fig:showcase}b) represents the KG for the denseblock4 layer of DenseNet121 on the Husky vs. Wolf dataset. This graph reveals three concepts -- ``forest'', ``mountain'', and ``wood'' -- as model-biased, marked in red edges. These biases imply that DenseNet121 associates these background treats with a specific class (e.g., Wolf) potentially due to overfitting to environmental contexts in the training data.

Those examples highlight misalignments discovered using BAGEL. Its interactive nature further supports a detailed exploration of concept-class dynamics.

\begin{figure}[htpb]
\centering
\includegraphics[width=\linewidth]{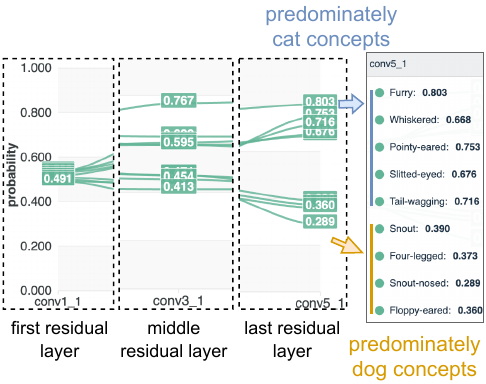}
\caption{\textbf{Example of concept detection dynamics across the layers.} Probabilities show if the concept is present in VGG's residual layer features for a class Cat upon Cats/Dogs dataset. }
\label{fig:concept_layer_dynamic}
\end{figure}

In addition, we investigate the transition of the concept probabilities within the layers embeddings. Figure \ref{fig:concept_layer_dynamic} shows the evolution of these probabilities across the layers of the VGG model trained on the Cats/Dogs dataset. Concept probabilities are derived using logistic regression applied to the embeddings, as described in Section \ref{sec:concept_prediction_using_embeddings}. Initially, the early layers tend to mix concepts with limited relevance to the specific class. As the network deepens, mid-level layers begin to filter and emphasize concepts more aligned with the input. In the final layer, the model typically enhances the probabilities of relevant concepts while suppressing those deemed irrelevant. The plot is auto-generated alongside the knowledge graph (KG) and is also available on our demo website for all experimental setups (models and datasets).

\begin{figure}[htpb]
\centering
\includegraphics[width=1.1\linewidth]{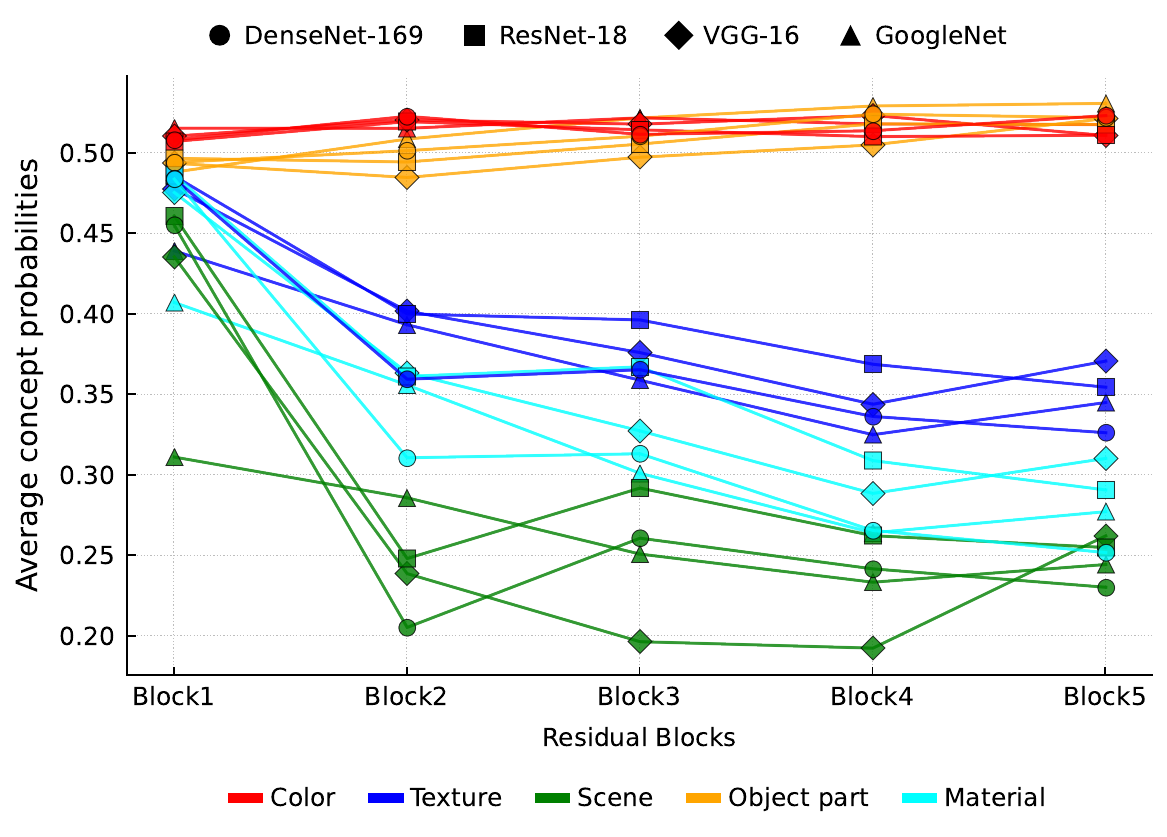}
\caption{Dynamics of concept representation across layers. Colors and textures are primarily detected in earlier layers, while object parts are more prominent in deeper layers.} 
\label{fig:concepts_layer_dynamic}
\end{figure}

Drawing on the methodologies of \cite{netdissect2017, bau2020understanding} and \cite{tcav2020}, we also investigate the general tendency of the emergence of conceptual categories across the residual layers (Figure \ref{fig:concepts_layer_dynamic}). Expectedly, rudimentary concepts, e.g., materials and textures, consistently manifest in the initial layers. In contrast, deeper layers progressively demonstrate higher probabilities of more complex concepts being present, such as object parts.

\subsection{Quantitative results}
\label{sec:comparison}

While XAI techniques can successfully identify concepts present in the model's learned representations, the detection of them does not inherently confirm that the model applies it causally in its decision-making, as feature attribution methods may capture correlations rather than mechanistic dependencies \cite{adebayo2018sanity}.

To evaluate the effectiveness of our proposed method, we conduct a comparison with two approaches: Testing with Concept Activation Vectors (TCAV) \cite{tcav2020} and Sparse Autoencoder (SAE) \cite{karvonen2024sae}. These baselines are described in Supp. \ref{appndx:comparison}. The results, summarized in Table~\ref{tab:method_comparison}, report the recall scores for concept detection across different network blocks. The recall is calculated by comparing the top 5 concepts biased towards one class in the dataset, as described in Section \ref{sec:Dataset}, with the top-$k$ biased concepts detected by the techniques, measuring the proportion of true positive concepts correctly identified. 

\begin{table}[htpb]
\centering
\caption{Comparison of recall scores for biased concept detection across network architectures and blocks for TCAV, SAE, and BAGEL (our method). Best block performance and average block scores are included. Higher values indicate better biased concept detection.}
\label{tab:method_comparison_horizontal}
    \resizebox{0.5\textwidth}{!}
    {
\begin{tabular}{l|l l l}
\toprule
\textbf{Method} & \textbf{Cats vs Dogs} & \textbf{MonuMAI} & \textbf{KitFox vs RedFox} 
\\
\midrule
TCAV & \textbf{0.900} & \textbf{0.885} & \textbf{1.000}\\
SAE & 0.450 & 0.650 & 0.950 \\
BAGEL JS & 0.350 & 0.883 & \textbf{1.000}\\
BAGEL F1 & 0.800 & 0.800 & 0.500 \\
\midrule
\textbf{Method}  & \textbf{Husky vs Wolf} & \textbf{Derm7pt} \\
\midrule
TCAV  & 0.600 & 0.575 \\
SAE  & 0.275 & 0.450 \\
BAGEL Jensen & 0.550 & 0.700 \\
BAGEL F1 & \textbf{0.875} & \textbf{0.800} \\
\bottomrule
\end{tabular}
}
\end{table}


\textbf{Discussion on the results.} We emphasize that properly evaluating the quality of an XAI technique is a challenging task. In this work, we propose to assess the quality of an XAI method based on its ability to detect dataset biases that might influence the model's decisions. While there is no guarantee that such biases indeed affect the model, we use recall as a metric to evaluate whether the prominent biased concepts in the dataset are also identified by the explanation method. It is important to note that the model may exhibit additional biases beyond those present in the dataset. Our goal is not to claim that achieving the highest recall necessarily indicates the best explanation method. Rather, we argue that an XAI technique should be able to recover at least some of the dataset’s underlying biases. From this perspective, \textbf{BAGEL} often demonstrates strong and consistent performance, outperforming \textbf{TCAV}, which tends to be less stable, and \textbf{SAE}, which generally performs poorly.

%% file: sections/conclusion.tex
In this paper, we have introduced \textbf{BAGEL}, a scalable framework for the global dissection of concept-based biases in DNNs.
Unlike traditional interpretability methods that focus on local predictions or isolated neurons, BAGEL offers a global perspective by modeling how high-level semantic concepts emerge, interact, and propagate across the internal layers of a network. Our method bridges the gap between dataset-driven and model-learned concept representations, enabling systematic analysis of alignment or divergence between them.

By visualizing these relationships through a structured knowledge graph, BAGEL empowers users to uncover spurious correlations, assess model biases, and gain deeper insight into the decision-making processes of complex models. Importantly, our approach is model-agnostic, scalable, and complements existing interpretability efforts by focusing not only on what a model predicts, but how and why it does so.

Through this work, we take a step toward more transparent, accountable, and trustworthy AI systems by providing both theoretical tools and practical interfaces for analyzing model behavior at a semantic level. Future work will explore extending BAGEL to multimodal architectures and refining concept definitions through automated discovery.

\paragraph{Acknowledgments.}
 This work was performed using HPC resources from GENCI-IDRIS (Grant 2024-[AD011011970R4] and Grant 2024-[AD011015965]).

%% file: sections/appendix.tex
\appendix

\startcontents
\renewcommand\contentsname{Table of Contents - Supplementary Material}
{
\hypersetup{linkcolor=black}
\printcontents{ }{1}{\section*{\contentsname}}{}
}

\clearpage

\input{si/si1}

\input{si/si2}

\input{si/si3}

%% file: si/si1.tex
\section{Extra results}
\label{appndx:comparison}

To build our baseline, we compare \textbf{BAGEL} with two alternative approaches: Testing with Concept Activation Vectors (TCAV) and Sparse Autoencoder (SAE). Although these methods are not specifically designed for the exact same task, we adapt them to our setting for comparison.

For \textbf{TCAV}, we use the concept importance scores it provides to determine whether a given concept is relevant in the classification process. TCAV is applied at each layer of the DNN, and we report only the best-performing results across layers.

For the \textbf{SAE}-based approach, we train a sparse autoencoder and, for each latent representation, identify the set of concepts associated with it. This is done by assigning, for each image, the most activated coefficient from the SAE representation. Each coefficient is then associated with a set of images, and since each image is annotated with a set of concepts, we assign to each coefficient the most prominent concept across its image set. This allows us to identify which concepts are most influential for a given layer of a DNN when classifying an image.

Using \textbf{BAGEL}, \textbf{SAE}, and \textbf{TCAV}, we extract the most prominent concepts responsible for a given classification. We then compute a recall metric by selecting the top-$k$ most prominent concepts per instance and comparing them to the top-5 most frequent biased concepts at the dataset level.

It is important to note that while these methods highlight the importance of certain concepts for a DNN’s decision, they do not necessarily indicate whether the DNN is truly relying on biased information.

\begin{table*}[htpb]
\centering
\caption{Comparison of recall scores for biased concept detection across network architectures and blocks for TCAV, SAE, and our method. Best block performance and average block scores are included. Higher values indicate better biased concept detection.}
\label{tab:method_comparison}
\resizebox{\textwidth}{!}{
\begin{tabular}{ll *{8}{S[table-format=1.2]}}
\toprule
\multicolumn{10}{c}{\textbf{Architecture}} \\
\cmidrule(lr){3-10}
& & {DenseNet121} & {VGG16} & {DenseNet169} & {EfficientNet} & {ResNeXt101} & {AlexNet} & {GoogLeNet} & {ResNet18} \\
\midrule
\multicolumn{10}{c}{\textbf{HuskyVsWolf}} \\
\midrule
\multirow{2}{*}{TCAV}       & Best Block & 0.60 & 1.00 & 0.80 & 0.40 & 0.60 & 0.40 & 0.60 & 0.40 \\
                            & Avg. Block & 0.32 & 0.60 & 0.52 & 0.16 & 0.28 & 0.32 & 0.44 & 0.24 \\
\multirow{2}{*}{SAE}        & Best Block & 0.40 & 0.00 & 0.40 & 0.20 & 0.20 & 0.20 & 0.40 & 0.40 \\
                            & Avg. Block & 0.24 & 0.00 & 0.28 & 0.12 & 0.12 & 0.04 & 0.20 & 0.12 \\
\multirow{2}{*}{BAGEL F1 (Ours)}   & Best Block & 0.80 & 1.00 & 0.80 & 1.00 & 0.80 & 1.00 & 0.80 & 0.80 \\
                                   & Avg. Block & 0.68 & 0.68 & 0.68 & 0.80 & 0.76 & 0.76 & 0.76 & 0.72 \\
\multirow{2}{*}{BAGEL JS (Ours)}   & Best Block & 0.40 & 0.60 & 0.40 & 0.80 & 0.60 & 0.60 & 0.60 & 0.40 \\
                                       & Avg. Block & 0.40 & 0.52 & 0.40 & 0.48 & 0.44 & 0.56 & 0.48 & 0.40 \\
\midrule
\multicolumn{10}{c}{\textbf{Cats Dogs}} \\
\midrule
\multirow{2}{*}{TCAV}       & Best Block & 1.00 & 0.80 & 1.00 & 1.00 & 0.80 & 0.80 & 0.80 & 1.00 \\
                            & Avg. Block & 0.76 & 0.76 & 0.92 & 0.80 & 0.76 & 0.72 & 0.68 & 0.80 \\
\multirow{2}{*}{SAE}        & Best Block & 0.60 & 0.40 & 0.60 & 0.40 & 0.40 & 0.40 & 0.40 & 0.40 \\
                            & Avg. Block & 0.40 & 0.24 & 0.44 & 0.28 & 0.24 & 0.28 & 0.28 & 0.28 \\
\multirow{2}{*}{BAGEL F1 (Ours)}   & Best Block & 0.80 & 0.80 & 0.80 & 0.80 & 0.80 & 0.80 & 0.80 & 0.80 \\
                                   & Avg. Block & 0.72 & 0.72 & 0.76 & 0.64 & 0.72 & 0.64 & 0.64 & 0.64 \\
\multirow{2}{*}{BAGEL JS (Ours)}   & Best Block & 0.40 & 0.20 & 0.40 & 0.40 & 0.40 & 0.40 & 0.40 & 0.20 \\
                                   & Avg. Block & 0.24 & 0.20 & 0.24 & 0.24 & 0.28 & 0.28 & 0.24 & 0.20 \\
\midrule
\multicolumn{10}{c}{\textbf{Derm7pt}} \\
\midrule
\multirow{2}{*}{TCAV}       & Best Block & 0.60 & 0.40 & 0.60 & 0.60 & 0.80 & 0.40 & 0.60 & 0.60 \\
                            & Avg. Block & 0.44 & 0.40 & 0.60 & 0.44 & 0.80 & 0.40 & 0.60 & 0.60 \\
\multirow{2}{*}{SAE}        & Best Block & 0.60 & 0.40 & 0.60 & 0.40 & 0.40 & 0.40 & 0.40 & 0.40 \\
                            & Avg. Block & 0.44 & 0.28 & 0.44 & 0.32 & 0.24 & 0.28 & 0.20 & 0.32 \\
\multirow{2}{*}{BAGEL F1 (Ours)}   & Best Block & 0.80 & 0.80 & 0.80 & 0.80 & 0.80 & 0.80 & 0.80 & 0.80 \\
                                   & Avg. Block & 0.64 & 0.64 & 0.68 & 0.64 & 0.72 & 0.64 & 0.64 & 0.56 \\
\multirow{2}{*}{BAGEL JS (Ours)}   & Best Block & 0.80 & 0.60 & 0.60 & 0.60 & 0.80 & 0.60 & 0.80 & 0.80 \\
                                   & Avg. Block & 0.68 & 0.60 & 0.60 & 0.60 & 0.72 & 0.52 & 0.64 & 0.68 \\
\midrule
\multicolumn{10}{c}{\textbf{KitFoxVsRedFox}} \\
\midrule
\multirow{2}{*}{TCAV}       & Best Block & 1.00 & 1.00 & 1.00 & 1.00 & 1.00 & 1.00 & 1.00 & 1.00 \\
                            & Avg. Block & 1.00 & 1.00 & 1.00 & 1.00 & 1.00 & 1.00 & 1.00 & 1.00 \\
\multirow{2}{*}{SAE}        & Best Block & 1.00 & 1.00 & 1.00 & 1.00 & 1.00 & 0.80 & 0.80 & 1.00 \\
                            & Avg. Block & 0.88 & 0.84 & 0.88 & 0.84 & 0.84 & 0.80 & 0.80 & 0.88 \\
\multirow{2}{*}{BAGEL F1 (Ours)}   & Best Block & 0.60 & 0.60 & 0.60 & 0.60 & 0.60 & 0.60 & 0.60 & 0.60 \\
                                   & Avg. Block & 0.60 & 0.60 & 0.60 & 0.60 & 0.60 & 0.60 & 0.60 & 0.60 \\
\multirow{2}{*}{BAGEL JS (Ours)}   & Best Block & 1.00 & 1.00 & 1.00 & 1.00 & 1.00 & 1.00 & 1.00 & 1.00 \\
                                   & Avg. Block & 1.00 & 1.00 & 1.00 & 1.00 & 1.00 & 1.00 & 1.00 & 1.00 \\
\midrule
\multicolumn{10}{c}{\textbf{MonuMAI}} \\
\multirow{2}{*}{TCAV}       
& Best Block & 0.80 & 0.80 & 0.80 & 1.00 & 1.00 & 0.80 & 0.60 & 1.00 \\
& Avg. Block  & 0.64 & 0.64 & 0.52 & 0.80 & 0.72 & 0.52 & 0.52 & 0.68 \\
\multirow{2}{*}{SAE}        
& Best Block & 0.80 & 0.60 & 0.60 & 0.60 & 0.80 & 0.80 & 0.60 & 0.60 \\
& Avg. Block  & 0.60 & 0.56 & 0.52 & 0.48 & 0.52 & 0.56 & 0.56 & 0.56 \\
\multirow{2}{*}{BAGEL JS (Ours)} 
& Best Block & 1.00 & 0.80 & 1.00 & 1.00 & 1.00 & 0.80 & 1.00 & 1.00 \\
& Avg. Block  & 0.80 & 0.68 & 0.76 & 0.84 & 0.84 & 0.76 & 0.84 & 0.92 \\
\multirow{2}{*}{BAGEL F1 (Ours)} 
& Best Block & 0.80 & 0.80 & 0.80 & 0.80 & 0.80 & 0.80 & 0.80 & 0.80 \\
& Avg. Block  & 0.80 & 0.80 & 0.80 & 0.80 & 0.80 & 0.80 & 0.80 & 0.80 \\
\bottomrule
\end{tabular}
}
\end{table*}

%% file: si/si2.tex
\section{Training details}
\subsection{Training hyperparameters}
\label{appndx:training_details}

On each of the tested datasets, the models were initialized with the pre-trained ImageNet weights and trained until the validation loss was not improving of at least 0.001 for 5 consequential epochs. The batch size of 64 was used for all trainings, as well as a momentum of 0.9 for the SGD optimizer. The other used hyperparameters are shown in the table. 

\begin{table}[htbp]
\centering
\caption{Hyperparameters used for fine-tuning each model.}
\begin{tabular}{lccc}
\toprule
\textbf{Model} & \textbf{Learning Rate} & \textbf{Weight Decay} & \textbf{Optimizer} \\
\midrule
AlexNet        & $1 \times 10^{-4}$ & $5 \times 10^{-4}$ & Adam     \\
DenseNet121    & $5 \times 10^{-5}$ & $1 \times 10^{-4}$ & SGD      \\
DenseNet169    & $5 \times 10^{-5}$ & $1 \times 10^{-4}$ & SGD      \\
EfficientNet-B0   & $3 \times 10^{-5}$ & $1 \times 10^{-5}$ & AdamW    \\
GoogLeNet    & $1 \times 10^{-4}$ & $2 \times 10^{-4}$ & Adam     \\
ResNet18       & $1 \times 10^{-3}$ & $1 \times 10^{-4}$ & SGD      \\
ResNeXt101     & $5 \times 10^{-4}$ & $1 \times 10^{-4}$ & SGD      \\
VGG16          & $1 \times 10^{-4}$ & $5 \times 10^{-4}$ & SGD      \\
\bottomrule
\end{tabular}
\label{tab:hyperparams}
\end{table}

%% file: si/si3.tex
\subsection{Set of concepts}
\label{appndx:set_of_concepts}

\begin{table}[htbp]
\centering
\caption{Datasets and their set of concepts grouped by class.}
\label{tab:datasets_and_concepts}
\begin{tabular}{llp{9cm}}
\toprule
\textbf{Dataset} & \textbf{Class} & \textbf{Concepts} \\
\midrule
\multirow{4}{*}{MonuMAI} 
  & Baroque     & arco-conopial, columna-salomonica, fronton-curvo \\
  & Gothic      & arco-apuntado, pinaculo-gotico \\
  & Islamic     & arco-herradura, arco-lobulado, arco-trilobulado, dintel-adovelado \\
  & Renaissance & arco-medio-punto, fronton, fronton-partido, serliana, vano-adintelado, ojo-de-buey \\
\midrule
\multirow{2}{*}{Kit Fox vs Red Fox} 
  & Kit Fox & Large-ears, Small-body, Slim-build, Bushy-tail, Black-tipped-tail, Tan-fur, Dark-muzzle, Short-legs, White-underside, Delicate-face \\
  & Red Fox & Reddish-fur, White-underside, Black-stockings, White-tipped-tail, Dark-Ear-Rims, Narrow-Snout, Slender-Body, Long-Legs, Amber-Eyes, Fluffy-Fur \\
\midrule
\multirow{2}{*}{Husky vs Wolf} 
  & Husky & Thick double coat, Almond-shaped eyes, Erect triangular ears, Distinctive facial markings, Bushy tail, Medium to large size, Athletic build, Variety of coat colors, White fur on paws, White fur on chest \\
  & Wolf  & Large size, Long legs, Bushy tail, Pointed ears, Varied coat colors, Yellow eyes, Broad chest, Long muzzle, Large paws, Thick fur \\
\midrule
\multirow{2}{*}{Cats, Dogs} 
  & Cat & Furry, Whiskered, Pointy-eared, Slitted-eyed, Four-legged \\
  & Dog & Snout, Wagging-tailed, Snout-nosed, Floppy-eared, Tail-wagging \\
\midrule
\multirow{7}{*}{Derm7pt}
  & Pigment Network & typical pigment network, atypical pigment network \\
  & Streaks & irregular streaks, regular streaks \\
  & Pigmentation & diffuse irregular pigmentation, localized irregular pigmentation, diffuse regular pigmentation, localized regular pigmentation \\
  & Regression Structures & blue areas regression structures, combinations regression structures, white areas regression structures \\
  & Dots and Globules & irregular dots and globules, regular dots and globules \\
  & Blue Whitish Veil & blue whitish veil \\
  & Vascular Structures & arborizing vascular structures, within regression vascular structures, hairpin vascular structures, dotted vascular structures, comma vascular structures, linear irregular vascular structures, wreath vascular structures \\
\bottomrule
\end{tabular}
\end{table}